\documentclass[conference]{IEEEtran}
\IEEEoverridecommandlockouts
\usepackage{cite}
\usepackage{amsmath,amssymb,amsfonts}
\usepackage{graphicx}
\usepackage{textcomp}
\usepackage{xcolor}
\usepackage{algorithm}
\usepackage{adjustbox}
\usepackage{multirow}
\usepackage{array}
\usepackage{nicefrac}
\usepackage{algpseudocode}
\def\BibTeX{{\rm B\kern-.05em{\sc i\kern-.025em b}\kern-.08em
    T\kern-.1667em\lower.7ex\hbox{E}\kern-.125emX}}

\definecolor{darkspringgreen}{rgb}{0.09, 0.45, 0.27}

\newcommand{\StatexIndent}[1][3]{%
  \setlength\@tempdima{\algorithmicindent}%
  \Statex\hskip\dimexpr#1\@tempdima\relax}

\usepackage{tikz}
\usepackage{textcomp}
\usepackage[doipre={DOI:~}]{uri}
\usepackage{lipsum}
\newcommand\copyrighttext{%
  \footnotesize 
  \textbf{Accepted for publication at the ISLPED 2022 ACM/IEEE International Symposium on Low Power Electronics and Design}.
  }
\newcommand{\copyrightnotice}{%
\begin{tikzpicture}[remember picture,overlay]
\node[anchor=south,yshift=10pt] at (current page.south) {\fbox{\parbox{\dimexpr\textwidth-\fboxsep-\fboxrule\relax}{\copyrighttext}}};
\end{tikzpicture}%
}

\begin{document}
\bstctlcite{IEEEexample:BSTcontrol}

\title{Multi-Complexity-Loss DNAS for Energy-Efficient and Memory-Constrained Deep Neural Networks
}
\author{\IEEEauthorblockN{Matteo Risso\IEEEauthorrefmark{2}, Alessio Burrello\IEEEauthorrefmark{1}, Luca Benini\IEEEauthorrefmark{1}, Enrico Macii\IEEEauthorrefmark{3}, Massimo Poncino\IEEEauthorrefmark{2}, Daniele Jahier Pagliari\IEEEauthorrefmark{2}}
\IEEEauthorblockA{\IEEEauthorrefmark{1}Department of Electrical, Electronic and Information Engineering, University of Bologna, 40136 Bologna, Italy\\
\IEEEauthorrefmark{2}Department of Control and Computer Engineering, Politecnico di Torino, Turin, Italy\\
\IEEEauthorrefmark{3}Inter-university Department of Regional and Urban Studies and Planning, Politecnico di Torino, Turin, Italy\\
}
\IEEEauthorblockA{Corresponding Email: matteo.risso@polito.it}
\vspace{-1.1cm}
}

\maketitle

\copyrightnotice

\begin{abstract}
%
Neural Architecture Search (NAS) is increasingly popular to automatically explore the accuracy versus computational complexity trade-off of Deep Learning (DL) architectures. When targeting tiny edge devices, the main challenge for DL deployment is matching the tight memory constraints, hence most NAS algorithms consider model size as the complexity metric. Other methods reduce the energy or latency of DL models by trading off accuracy and number of inference operations. Energy and memory are rarely considered simultaneously, in particular by low-search-cost Differentiable NAS (DNAS) solutions.

We overcome this limitation proposing the first DNAS that directly addresses the most realistic scenario from a designer's perspective: the co-optimization of accuracy and energy (or latency) \textit{under a memory constraint}, determined by the target HW. We do so by combining two complexity-dependent loss functions during training, with independent strength. Testing on three edge-relevant tasks from the MLPerf Tiny benchmark suite, we obtain rich Pareto sets of architectures in the energy vs. accuracy space, with memory footprints constraints spanning from 75\% to 6.25\% of the baseline networks. When deployed on a commercial edge device, the STM NUCLEO-H743ZI2, our networks span a range of 2.18x  in energy consumption and 4.04\% in accuracy for the same memory constraint, and reduce energy by up to 2.2$\times$ with negligible accuracy drop with respect to the baseline.

%
%
%
%
%
\end{abstract}

\begin{IEEEkeywords}
Deep Learning, TinyML, Energy-efficiency, NAS
\end{IEEEkeywords}

\section{Introduction}
Deep Learning (DL) is at the core of many modern computing applications, such as computer vision~\cite{resnet}, sound classification~\cite{zhang2017hello}, bio-signal analysis~\cite{burrello2021q}, predictive maintenance~\cite{burrello2020predicting}, etc.
While DL models have been traditionally deployed on powerful cloud-based servers,
evidence exists about the potential advantages of an implementation at-the-edge~\cite{Daghero2021a}.
Edge computing could improve privacy and reduce the energy consumption at the distributed system level, by replacing the energy hungry wireless transmission of raw data with more efficient local computations and transmission of aggregated outputs~\cite{edge_computing_2016}.

This has spurred strong academic and industral interest for so-called \textit{TinyML}, i.e., the study of techniques and tools to enable the deployment of Machine Learning (ML) and DL models on low power, battery-operated edge devices.
In this context, the key hard-requirement to be satisfied is on the memory footprint of DL models, which should match the severe constraints of edge devices, typically based on Microcontrollers (MCUs) with few MBs of Flash and RAM~\cite{stm32h7}. At the same time, energy consumption should be minimized, typically by reducing the total number of operations (OPs) per prediction, in order to maximize the system's lifetime.

Achieving these goals through a manual tuning of a Deep Neural Network's (DNN) hyper-parameters, while maintaining a sufficient prediction accuracy, is a tedious and time-consuming process.
Therefore, Neural Architecture Search (NAS) tools have emerged as new design space exploration and automation solutions, able to find DNN hyper-parameters that co-optimize prediction performance and a computational cost metric, such as the number of parameters (i.e., \textit{Size}), the number of OPs per inference, or the latency/energy consumption~\cite{gordon2018morphnet,cai2018proxylessnas}. However, classic NASes, e.g., based on reinforcement learning, are extremely time-consuming (1000s of GPU hours), thus being inaccessible to most edge systems designers, while light-weight Differentiable NAS (DNAS) solutions are limited in the ways they can express optimization objectives and constraints. Indeed, to our knowledge, all existing DNAS methods optimize either the model size or the number of OPs \textit{separately}. In contrast, the relevant problem from a designer's perspective is the \textit{minimization of energy (OPs) under a given memory constraint}.
\begin{figure}[t]
  \centering
  \includegraphics[width=\columnwidth]{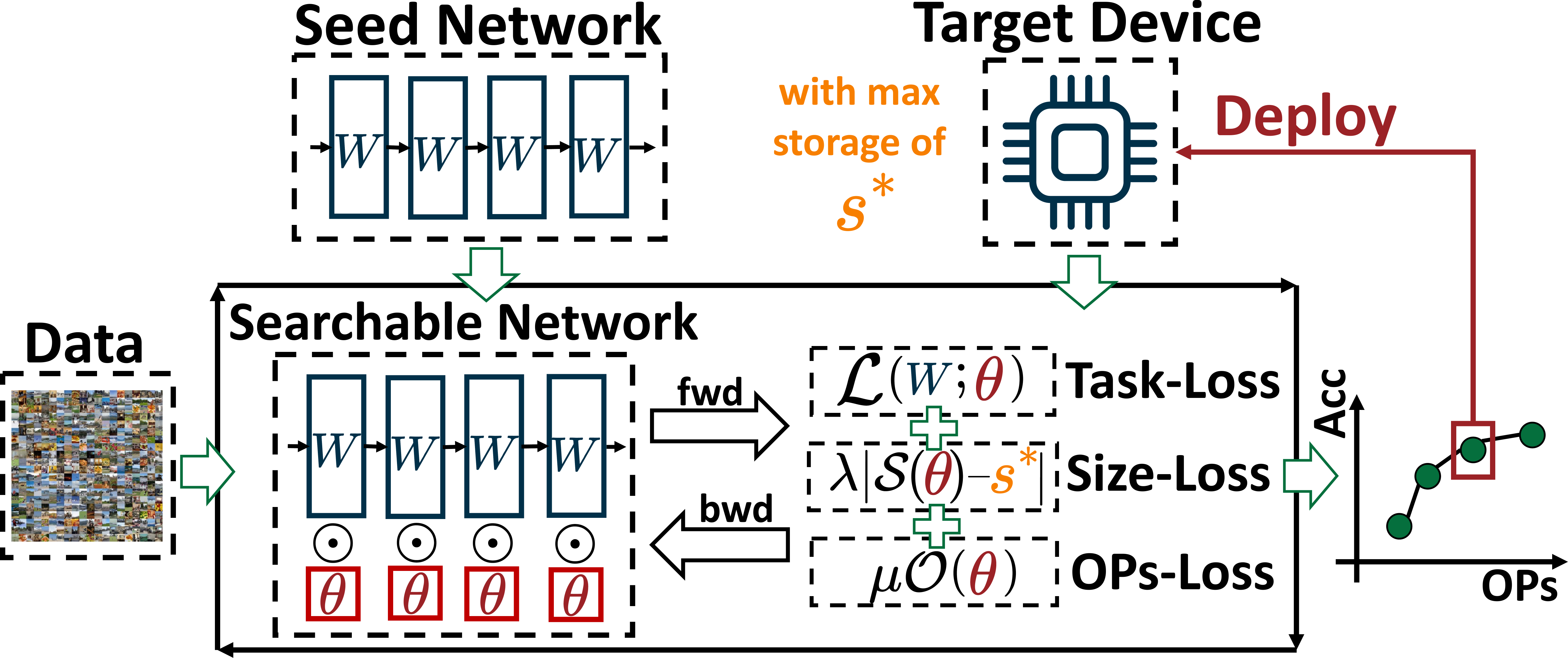}
  \vspace{-0.6cm}
  \caption{Overview of the proposed approach.}
  \label{fig:flow}
\end{figure}

In this work, we address this issue proposing a novel problem formulation (shown in Fig.~\ref{fig:flow}) that can be applied to any DNAS, and allows to find a set of Pareto-optimal architectures in the accuracy vs OPs space, under a fixed model size constraint.
We validate our formulation in combination with a simple DNAS, that performs a fine-grained search over the number of channels in Convolutional Neural Network (CNNs) layers.
With experiments on three different benchmarks from the TinyMLPerf suite~\cite{banbury2020benchmarking} we show that, starting from networks already optimized for edge deployment, our method can further improve and enrich the Pareto frontier.
Deploying the automatically discovered architectures on a real edge device, the STM NUCLEO-H743ZI2, we show that we can reduce the energy consumption up to 2.2$\times$ with negligible accuracy drop, while also cutting memory occupation, compared to the baseline hand-tuned CNNs. Our code is open-sourced at: \texttt{https://not-yet-available}.

\section{Background and Related Works}
Initial attempts to enable inference at-the-edge were based on hand-crafting efficient DNN architectures. Notable examples include SqueezeNet~\cite{iandola2016squeezenet}, MobileNets~\cite{howard2017mobilenets}, EfficientNet~\cite{tan2019efficientnet}, etc.
Such models, originally proposed for mobile deployment, are the result of a long and time-consuming manual tuning of hyper-parameters based on heuristics and human ingenuity. %
While very efficient, hand-tuned DNNs are never a one-size-fits-all: they represent a single point in the accuracy versus complexity space, which cannot fit all deployment scenarios. For instance, they cannot be directly applied in TinyML use cases, where hardware has much tighter constraints with respect to mobile systems. Accordingly, in order to prevent designers from having to repeat such hyperparameters' hand-tuning from scratch for each prediction task and deployment target, research has recently focused on automated DNN optimization solutions referred to as ``NAS''.

Early NAS algorithms explore the search space by means of Evolutionary Algorithms (EA)~\cite{nas_evol} or training a Reinforcement Learning (RL) agent~\cite{nas_reinforcement_2016, NASNet, mnasnet_2019}.
At each iteration they sample and \textit{fully train} one or more architectures from the search space.
The obtained accuracy and cost (e.g., model size, OPs, etc.) are then used to generate the EA fitness function or RL reward. 
Such solutions are extremely powerful, allowing to accommodate any combination of optimization target and constraints. However, they do not scale well with the dimension of the search space, requiring thousands of GPU hours for a single search, due to the repeated sequence of sampling, training, and evaluation. So-called ``proxies'', e.g., training for a reduced number of epochs or on a reduced dataset, can cut the search complexity, but may also undermine the quality of results~\cite{cai2018proxylessnas}.

Alternatively, search costs can be reduced resorting to more recent DNAS methods, which optimize a DNN architecture \textit{while training it}, using the same gradient descent algorithms to optimize both the weights and the network hyperparameters (e.g., depth, neurons, receptive field, etc).
One way to achieve this result is through the use of \textit{super-nets}, large DNNs including multiple alternative implementations of each layer/module, with different hyperparameters settings~\cite{liu2018darts}. Each super-net path corresponds to a potential final architecture, and the optimal one is selected during training, using a differentiable relaxation of the problem: alternative layers' are combined by means of a set of continuous and trainable \textit{architectural weights}, which are then optimized by gradient-descent, so to assign larger weights to the alternatives that maximize the metric(s) of interest.
At the end of the training, a discretization step selects a single path, typically the one associated with the largest architectural weights.

In DNASes, networks that are simultaneously accurate and low-complexity are typically found enhancing the standard, task-dependent, loss function $\mathcal{L}$ with an additional regularization term $\mathcal{R}$, that models the cost metric to be optimized (size, OPs, energy/latency, etc.) as a differentiable function. The overall optimization goal becomes:
\begin{equation} \label{eq:dnas}
\min_{W, \theta} \mathcal{L}(W; \theta) + \lambda \mathcal{R}(\theta)
\end{equation}
where $W$ is the set of trainable weights of the network (e.g., convolutional kernels), $\theta$ is the set of NAS architectural weights encoding the different paths in the super-net and $\lambda$ is a scalar regularization strength that controls the relative importance between the task-specific loss and complexity loss.

While super-net-based DNAS can find an optimized architecture with a single training, thus being much faster than RL/EA methods, such training is still tricky when dealing with large search spaces, due to the explosion of the super-net size, which causes huge training time and memory overheads with respect to a ``normal'' DNN.
ProxylessNAS~\cite{cai2018proxylessnas} tackles the memory problem by sampling at most two super-net paths for each batch of inputs.

Other methods, such as FbNetV2~\cite{wan2020fbnetv2}, MorphNet~\cite{gordon2018morphnet} and PIT~\cite{risso2021pruning} replace the super-net with a standard DNN with a unique path, usually denoted as \textit{seed network}.
The search space is formed by sub-architectures of the seed, obtained by reduction of its hyper-parameters.
%
%
In practice, this result is obtained \textit{masking} different slices of each layer's weights with binary parameters, so that the slices multiplied with a 0 are effectively eliminated from the layer. The continuous relaxation of the binary mask is then optimized, similarly to the architectural weights in a super-net DNAS, with the objective of reducing the network complexity, by eliminating unimportant parts of each layer (in that, this approach is similar to a \textit{structured pruning}~\cite{Daghero2021a}).
The usage of masks introduces a minimum overhead with respect to a normal training of the seed~\cite{risso2021pruning}, reducing the search time and memory requirements significantly compared to super-net approaches, and representing a further step towards lightweight NAS.

One drawback of mask-based DNAS is that the search-space definition is less flexible, since it can only include reduced variants of the seed.
Nevertheless, this is traded with a much more fine-grained search granularity, hardly reproducible with multiple paths in a super-net. For instance, considering a convolutional layer with 32 channels, a mask-based DNAS can easily explore all variants of the number of feature maps with a granularity of 1 (i.e., 31, 30, 29, and so on); doing the same with a super-net DNAS would require a huge network with 32 alternative versions just for that layer.

%

\section{Proposed Method}
DNAS tools that follow the formulation of (\ref{eq:dnas}) have two main limitations.
First, $\mathcal{R}$ models a \textit{single cost metric}, i.e., either the model size, the number of OPs, or a differentiable approximation of the latency or energy consumption, as a function of the DNN hyper-parameters~\cite{gordon2018morphnet,cai2018proxylessnas}. 
Second, cost is considered as an objective to minimize, rather than a constraint. While this is appropriate for some metrics (e.g., OPs, latency or energy), it is sub-optimal when considering memory occupation. In fact, most designers are interested in finding the ``best'' model (e.g., the most accurate or the best balance between accuracy and energy consumption) that \textit{fits a memory constraint}, given by the target hardware. Doing so with (\ref{eq:dnas}) requires repeating the DNAS multiple times, sweeping $\lambda$, until a model with appropriate memory footprint is found.

We propose a new DNAS formulation that addresses both problems, allowing to consider both memory occupation and other cost metrics (OPs, in our experiments) simultaneously, taking the former as a constraint and the latter as an objective. In practice, this enables the search for Pareto-Optimal architectures in the Accuracy vs OPs plane, \textit{around a fixed memory budget}. 
We apply the proposed method on top of a mask-based DNAS that explores the number of channels $C_{out}$ of Convolutional layers in a CNN. However, our formulation is agnostic of the specific search method, and can be applied to \textit{any} DNAS, including super-net based ones.

The rest of this section is organized as follows.
Sec.~\ref{sec:diff_search} describes the considered search space, while
Sec.~\ref{sec:multireg_loss} presents the multi-regularization loss approach at the core of our method.
Finally, Sec.~\ref{sec:training_proc} details the training algorithm.

\subsection{Differentiable Channels Search} \label{sec:diff_search}
\begin{figure}[t]
  \centering
  \includegraphics[width=.9\columnwidth]{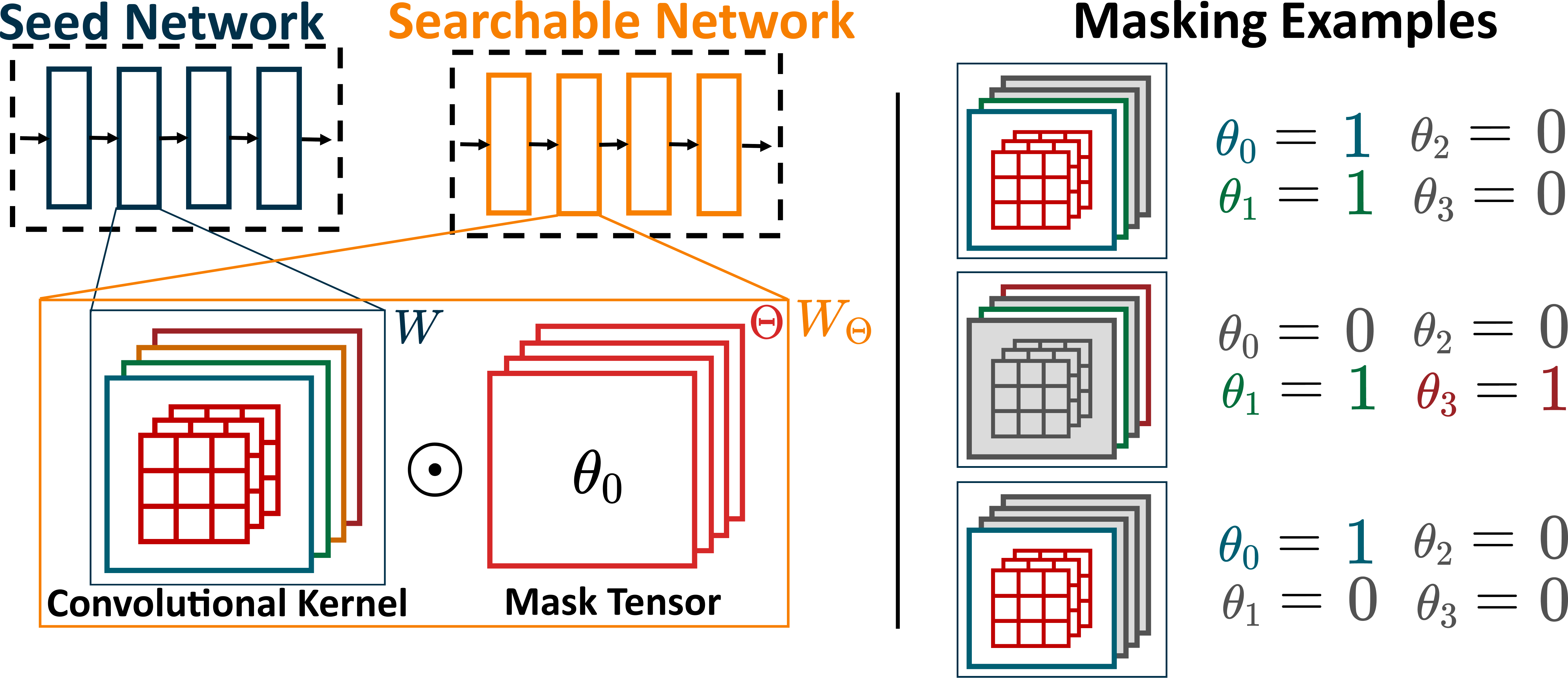}
  \vspace{-0.4cm}
  \caption{Proposed mask-based DNAS for convolutional layers output channels.}
  \label{fig:search_space}
\end{figure}
We apply our formulation to a simple but powerful mask-based DNAS, which optimizes the number of output channels of all convolutional kernels in a CNN seed.
This is a generalization of the approach proposed in~\cite{gordon2018morphnet}, which instead of adding explicit mask weights, made use of Batch Normalization (BN) parameters to eliminate some output channels. In contrast, our approach can work also when BN is not present, hence being applicable to \textit{any} CNN.

Fig.~\ref{fig:search_space} schematizes the proposed search scheme.
Starting from the seed network, a ``searchable'' model is built modifying the weight tensor $W^{(n)}$ of each convolutional layer, with size $C^{(n)}_{out} \times K^{(n)}_x \times K^{(n)}_y \times C^{(n)}_{in}$, where $C^{(n)}_{in}$, $C^{(n)}_{out}$ are the input/output channels and $K^{(n)}_x$/$K^{(n)}_y$ are the horizontal/vertical kernel sizes.
The searchable weights $W_{\Theta}^{(n)}$ are obtained as:
\begin{equation} \label{eq:masked_kernel}
W_{\Theta}^{(n)} = W^{(n)} \odot \mathcal{H}(\theta^{(n)})
\end{equation}
where $\theta^{(n)}$ is a trainable mask tensor with $C^{(n)}_{out}$ elements,
$\odot$ is the Hadamard product, and $\mathcal{H}$ is a Heaviside step function with a fixed threshold $th=0$, which has the effect of binarizing $\theta^{(n)}$ to 0/1 ($1 \text{ if } \theta \ge th, \text{ else } 0$). The product is \textit{broadcast} to an entire weight filter, i.e., the same element $\theta^{(n)}_i$ is multiplied with an entire slice $W^{(n)}_i$ of the weight tensor, of size $K^{(n)}_x \times K^{(n)}_y \times C^{(n)}_{in}$.
Therefore, the $i$-th element of the mask vector controls whether the $i$-th output channel is removed from the network ($\mathcal{H}(\theta^{(n)}_i) = 0$) or kept alive ($\mathcal{H}(\theta^{(n)}_i) = 1$).
The obtained ``searchable'' network is then inserted in a normal training loop, where $W$ and $\theta$ are trained together.

Specifically, in each forward pass of the training, Heaviside binarization has the effect of sampling a single architecture from the search-space (see the examples on the right of Fig.~\ref{fig:search_space}).
During backward passes, instead, a Straight-Through Estimator (STE) based on the BinaryConnect~\cite{courbariaux2015binaryconnect} approach lets gradients flow through the network despite the presence of the non-differentiable Heaviside Function.

\subsection{Multi-Regularization Loss} \label{sec:multireg_loss}
Most state-of-the-art DNAS tools~\cite{cai2018proxylessnas, gordon2018morphnet, wan2020fbnetv2} sum the task-specific loss $\mathcal{L}$ and the complexity term $\mathcal{R}$, scaled by a strength constant, using the scheme of (\ref{eq:dnas}).
%
%
MorphNet~\cite{gordon2018morphnet} and FBNetV2~\cite{wan2020fbnetv2} regularize either against the number of parameters or against the number of OPs, while ProxylessNAS~\cite{cai2018proxylessnas} tries to optimize the latency directly, using a model obtained fitting the latency measurements obtained profiling layers with different hyper-parameters combinations.
More recently, UDC~\cite{fedorov2022udc} proposed a different approach, where the regularization term includes a specific cost target $r^{*}$ to be satisfied, and the regularization term becomes $|\mathcal{R}(\theta) - r^{*}|$.

Building upon these ideas, we propose a novel formulation with \textit{two complexity loss terms}, which
%
%
drive
the DNAS towards a desired region of the search space, while still allowing the exploration of accuracy versus complexity trade-offs.
The proposed optimization problem formulation takes the form:
\begin{equation} \label{eq:multi_reg_eq}
\min_{W, \theta} \mathcal{L}(W; \theta) + \lambda |\mathcal{S}(\theta) - s^{*}| + \mu \mathcal{O}(\theta)
\end{equation}
In the equation, $\mathcal{S}$ models the size (i.e., memory footprint) of the DNN as a function of the architecture parameters $\theta$. In particular, for a CNN with N convolutional layers, $\mathcal{S}$ is computed as the total number of \textit{effective} (i.e., non-masked) parameters in those layers, i.e.:
\begin{equation} \label{eq:size_cost}
\mathcal{S}(\theta) = \sum_{n=0}^N \mathcal{S}^{(n)}(\theta) = \sum_{n=0}^N C_{out}^{(n-1)}(\theta) C_{out}^{(n)}(\theta)K_{x}^{(n)}K_{y}^{(n)}
\end{equation}
where $C_{out}^{(n-1)} = C_{in}^{(n)}$ and, for the 1st layer $C_{out}^{(n-1)}$ is fixed and equal to the number of channels in the input data.

Since, as explained above, memory occupation is usually a constraint that DNNs should respect for edge deployment, rather than a metric to optimize, we follow the approach of~\cite{fedorov2022udc}, minimizing the absolute value difference from a target size $s^{*}$ which depends on the hardware.

We associate this cost term with a relatively large and \textit{fixed} regularization strength $\lambda$, with $\lambda >> \mu$, thus forcing the NAS to find a set of $\theta^{*}$ parameters that yield $\mathcal{S}(\theta^{*}) \approx s^{*}$.
This allows us to immediately respect the memory constraint in each search, without a lengthy sweep of $\lambda$ values.
The way $\lambda$ is calculated for a given seed is detailed in Sec.~\ref{sec:training_proc}.

Furthermore, we add a further loss term $\mathcal{O}$ to model additional complexity-related metrics. In this work, $\mathcal{O}$ models the total number of OPs per prediction, which correlates with inference latency and energy consumption:
\begin{equation} \label{eq:ops_cost}
\mathcal{O}(\theta) = \sum_{n}^N \mathcal{S}^{(n)}(\theta) O^{(n)}_{x} O^{(n)}_{y}
\end{equation}
where $O^{(n)}_{x}$ and $O^{(n)}_{y}$ are respectively the output feature map width and height of the n-th convolutional layer.
%
%
We consider this metric as a general and easy-to-compute estimate of the inference cost of a model. However, our formulation is not limited to this specific expression for $\mathcal{O}$, and would be equally effective using more precise, profile-based energy or latency estimates, such as those proposed in~\cite{cai2018proxylessnas}.

Differently from $\mathcal{S}$, $\mathcal{O}$ is treated as an objective, not a constraint, and its importance is weighted by $\mu$,
which is the main knob used in our method to generate different final architectures starting from a single seed.
To summarize, the formulation of (\ref{eq:multi_reg_eq}) will produce DNNs with a size around $s^{*}$. With a small $\mu$, the DNAS will focus on minimizing $\mathcal{L}$, producing networks as accurate as possible (for that size constraint), while large $\mu$s will cause to partially sacrifice the accuracy in exchange for fewer OPs. 

Intuitively, since the OPs of a convolutional layer are equal to the parameters multiplied by the output feature map size (see Eq.~\ref{eq:ops_cost}), and since feature sizes tend to reduce going forward in the network, due to the effect of pooling, strided convolution, etc, OPs reduction under a fixed size budget can be obtained masking more channels in the \textit{initial} layers of the DNN, and less channels in the \textit{final} ones. We show that our formulation produces precisely this behavior in Sec.~\ref{sec:results}.

\subsection{Training Procedure} \label{sec:training_proc}
\begin{algorithm}[t]
\begin{algorithmic}[1]
\footnotesize
\caption{\label{alg:nas_search}}
    \For{$i \gets 1, \dots, \rm Epochs_{wu}$} {\color{darkspringgreen}\# warmup loop}
        \State Update $W$ based on $\nabla_{W} \mathcal{L}(W)$
    \EndFor
    \While{not converged} {\color{darkspringgreen}\# search loop}
        \State Update $W, \theta$ based on $\nabla_{W, \theta} (\mathcal{L}(W; \theta) + \lambda |\mathcal{S}(\theta) - s^{*}| + \mu \mathcal{O}(\theta) )$
    \EndWhile
    \For{$i \gets 1, \dots, \rm Epochs_{ft}$} {\color{darkspringgreen}\# fine-tuning loop}
        \State Update $W$ based on $\nabla_{W} \mathcal{L}(W)$
    \EndFor
\end{algorithmic}
\end{algorithm}
Alg.~\ref{alg:nas_search} shows the overall training scheme of our DNAS.
We start with a \textit{warmup phase}, i.e., a normal training of the full seed network, in which all masking parameters $\theta$ are frozen at the initialization value (i.e., 1), and only the normal weights $W$ are trained. The training objective in this phase consists solely of the task-specific loss function $\mathcal{L}$.
Noteworthy, warmup can be performed just once, saving the learned weights and reusing them for all following searches.

The second phase coincides with the actual architecture optimization.
In this step, the weights $W$ and the masking parameters $\theta$ are optimized together, to minimize the cumulative loss function of (\ref{eq:multi_reg_eq}).
The number of search epochs is controlled with an early-stop mechanism which monitors the task loss $\mathcal{L}$ on an unseen validation-split, and stops the search when the loss
stops improving.
%
%
%
When a validation set is not provided in the considered benchmark, we generate it by randomly sampling 10\% of the training set. 

Lastly, in the \textit{fine-tuning} phase, similarly to warmup, only the weights $W$ are trained against the task loss $\mathcal{L}$, while the $\theta$ architectural parameters are frozen to the final learned values.
In all our experiments, we set the warmup and fine-tuning epochs $\text{Epochs}_{\text{wu}}$ and $\text{Epochs}_{\text{ft}}$ equal to the number of training epochs used in the papers proposing each benchmark task. 

For a given target size $s^{*}$, the size strength $\lambda$ is determined with the formula $\lambda = \nicefrac{\mathcal{L}(\theta_{seed})}{|\mathcal{S}(\theta_{seed}) - s^{*}|}$, where $\mathcal{S}(\theta_{seed})$ and $\mathcal{L}(\theta_{seed})$ are the model size and task loss of the full seed network after warmup. The rationale is to have similar values for the first two addends of (\ref{eq:multi_reg_eq}) at the beginning of a search, so that the DNAS does not just shrink the network in the first iteration, ignoring completely the impact on accuracy. We found this heuristic to work well, but we also noticed that, as expected, varying $\lambda$ in a reasonable range ($\pm$ one order of magnitude) does not alter the search results significantly, since the term $\mathcal{S}(\theta) - s^{*}$ is quickly brought close to zero in the search phase. Importantly, this means that $\lambda$ can be computed in closed-form and does not have to be swept.

Having fixed the target size (and $\lambda$), multiple iterations of Alg.~\ref{alg:nas_search} with different values of $\mu$ generate a front of Pareto-optimal architectures in the Accuracy versus OPs space. 
Specifically, we always run a first search with $\mu=0$, to find the most accurate network which satisfies the $s^{*}$ constraint, without taking into account the number of OPs. We then progressively increase $\mu$ to find less accurate and more efficient architectures.
%
%
Importantly, too large $\mu$ values (violating $\lambda >> \mu$) lead to low-quality results, since the DNAS tries to dramatically reduce the number of OPs while simultaneously keeping the model size close to the target. This produces sub-optimal DNN architectures, with worse accuracy than those of smaller size. Thus, whenever increasing $\mu$ with fixed $s^{*}$ degrades the accuracy too much (in our experiments, we limit to a 5\% degradation w.r.t. the case $\mu=0$), we simply stop the exploration and switch to a lower size target.





\section{Experimental Results}\label{sec:results}
\begin{figure*}[t]
  \centering
  \vspace{-0.4cm}
  \includegraphics[width=0.9\textwidth]{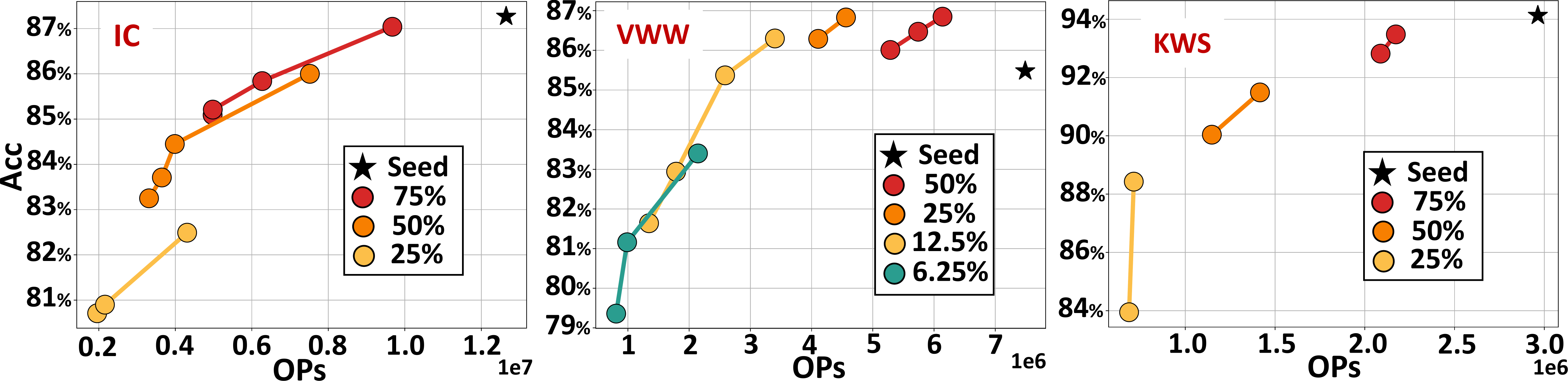}
  \vspace{-0.4cm}
  \caption{Accuracy versus OPs results for different size targets.}
  \label{fig:pareto_fronts}
\end{figure*}

\subsection{Setup}
We evaluated the proposed NAS on three datasets taken from the MLPerf Tiny Benchmark Suite~\cite{banbury2020benchmarking}. As seed networks, we used the reference architectures proposed in the suite for each dataset.
%
%
The \textit{Image Classification} (IC) benchmark is based on the well-known CIFAR-10 dataset,
%
%
which consists of 60000 32x32x3 RGB images belonging to 10 classes.
The reference CNN is a customized ResNet~\cite{resnet} with 8 convolutional layers.
The \textit{Visual Wake Word} (VWW) task considers the MSCOCO 2014 dataset,
%
%
with 109619 96x96x3 RGB images, and the objective is detecting whether at least a person is present in the input.
The reference architecture is a MobileNetV1~\cite{howard2017mobilenets} with a width multiplier of 0.25.
Lastly, the \textit{KeyWord Spotting} (KWS) benchmark is based on the Speech Commands v2 dataset,
%
%
which contains 105,829 utterances, to be classified in 12 classes including 10 words and two special labels (``unknown" and ``silence").
The reference architecture is the Depthwise Separable CNN (DS-CNN) described in~\cite{zhang2017hello}.
MLPerf Tiny includes a fourth Anomaly Detection benchmark. However, the reference DNN is an Autoencoder composed only of Dense layers, for which the model size and number of OPs are directly proportional (i.e., there is no degree of freedom to reduce the OPs under a fixed size budget). So, we did not consider that task, since it would not benefit from our formulation, which would become equivalent to (\ref{eq:dnas}).
%

Our DNAS is implemented in PyTorch v1.10.2.
Some relevant architectures found by our tool are then deployed on a commercial edge device, the NUCLEO-H743ZI2, in order to estimate energy consumption. We convert ONNX graphs exported from PyTorch into C code using the proprietary X-Cube-AI toolchain of STM.
In this work, we deploy floating-point models, but note that integer quantization is fully-orthogonal to our method. All results are reported on test sets.
\subsection{Search-Space Exploration}
Fig.~\ref{fig:pareto_fronts} shows the results obtained applying the proposed DNAS on the three benchmarks.
Each plot reports the found architectures (represented with coloured dots) and the seed (represented with a black star) in the Accuracy versus OPs space.
Different colors correspond to  different size targets $s^{*}$.
To validate our approach, we initially set $s^{*}$ to be respectively 75\%, 50\% and 25\% of the original size of each seed network. In a real scenario, $s^{*}$ would depend on the hardware, so this setup simulates targeting three different MCUs with progressively less available memory.
Within the curve relative to each $s^{*}$ target, different points are obtained changing the OPs regularization strength $\mu$.

The left-most graph shows the results obtained on the IC task.
Considering all three memory targets, the DNAS is able to find networks that span almost one order of magnitude in OPs 1.96M-9.86M), and $\pm$ 6.3\% in accuracy.
Moreover, under the 75\% size constraint, we obtain a network that achieves a negligible accuracy drop with respect to the seed (-0.23\%), while reducing the number of OPs by 1.3x (12.7M vs 9.86M).

The center graph reports the Pareto fronts obtained for the VWW task.
In this case, we found that the results obtained with the 75\% and 50\% size constraints are completely outperformed by those obtained with lower memory, which achieve higher accuracy with fewer OPs (only the 50\% curve is shown in the graph, for clarity). This means that the reference network used for this task is strongly over-parameterized.
Therefore, forcing to optimize OPs with a too high $s^{*}$, leads to un-balanced architectures which attain same accuracy of smaller ones (see orange vs. red curves in mid graph of Fig. \ref{fig:pareto_fronts}).
Thus, we decided to add two additional memory targets (i.e., 12.5\% and 6.25\% of the seed) in order to show more insightful trade-offs.
The results demonstrate once again that our DNAS is able to find a rich collection of Pareto-optimal architectures for multiple memory constraints.
The NAS results span almost one order of magnitude in terms of OPs (0.81M-6.14M).
Furthermore, many of the found architectures Pareto-dominate the seed, even at 12.5\% size (+0.39\% accuracy with 2.5$\times$ OPs reduction and +0.82\% accuracy with 2.2$\times$ OPs reduction).

Lastly, the right-most plot in Fig.~\ref{fig:pareto_fronts} shows the results on the KWS task.
In this case, Pareto-fronts are not as rich as for the other two benchmarks, due to the peculiarities of the seed network.
In fact, DS-CNN includes strided convolutions and pooling only in the first and last convolutional layers. Consequently, all intermediate feature map sizes are identical, with OPs and model size strongly correlated. 
Nonetheless, we still find multiple networks for each size constraint, although the trade-off between OPs and accuracy is less favorable. At most, for the 50\% size target, we obtain an OPs difference of 1.2$\times$ in exchange for an accuracy degradation of 1.45\%.

\subsection{Architecture Details}

\begin{figure}[t]
  \centering
  \vspace{-0.2cm}
  \includegraphics[width=\columnwidth]{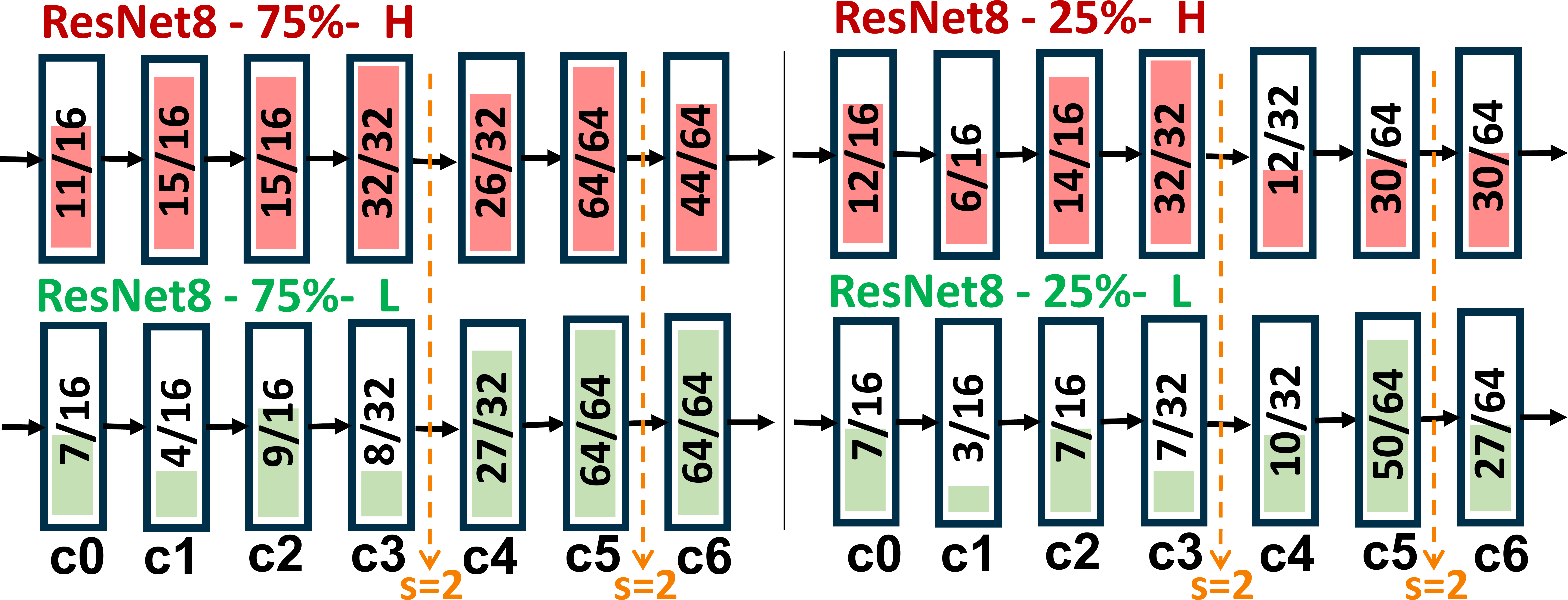}
  \vspace{-0.6cm}
  \caption{Examples of found architectures for the IC benchmark. }
  \label{fig:found_arch}
\end{figure}

As an example of the architectures found by our DNAS, Fig.~\ref{fig:found_arch} reports four of the networks generated for the IC benchmark, under the 75\% and 25\% size constraints.
The models labeled with ``-H" (high-OPs) are obtained with $\mu=0$, i.e., performing the search with the only constraint of respecting the target size, without OPs reduction.
Instead, the networks labeled with ``-L" (low-OPs) are obtained with with $\mu \ne 0$, and in particular, they correspond to the points with fewest OPs in the 75\% and 25\% Pareto fronts of the graph in Fig.~\ref{fig:pareto_fronts}. Each rectangle represents a convolutional layer, and numbers inside them correspond to $C^{(n)}_{out,final}/C^{(n)}_{out,seed}$.

These examples demonstrate that our formulation generates meaningful results. First, as expected, a lower target size results in more masked channels, regardless of the OPs regularization strength. Moreover, the ``-L'' networks have fewer channels in their \textit{initial} layers, which are those that contribute more to the total OPs, due to the larger resolution of their input/output feature maps.
The considered ResNet8 has two convolutional layers with stride $s=2$ (c3 and c5), indicated by yellow dashed lines in Fig.~\ref{fig:found_arch}, which reduce the feature map sizes of downstream layers by a factor 4. As evident from the figure, our DNAS reduces much more aggressively the layers before c3 when $\mu$ increases.

\subsection{Embedded Deployment}
\begin{table}[t]
\centering
\footnotesize
\vspace{-0.4cm}
\caption{Detailed deployment results for the IC and VWW benchmarks. The Mem. field express weights memory in kB.}\label{tab:deployment}
\vspace{-0.2cm}
\begin{adjustbox}{max width=1\columnwidth}
\begin{tabular}{|c|c|c|c|c|c|c|}
\cline{4-5}
\hline
 & & & Mem. & Lat. & En. \\
Task & Network & Acc. & [kB] (\% $\ne$ constr.) & [ms] & [mJ]\\\hline
\multirow{6}{*}{IC} & Seed & 87.27\% & 310 & 125 & 29.3\\
& 75\%-H & 87.04\% & 231.6 (-0.37\%) & 110 & 25.7\\
& 75\%-L & 85.09\% & 233.5 (+0.44\%) & 57.4 & 13.4\\
& 50\%-H & 86.00\% & 155.8 (+0.54\%) & 87.7 & 20.5\\
& 50\%-L & 83.22\% & 156 (+0.65\%) & 42.1 & 9.84\\
& 25\%-H & 82.49\% & 74.96 (-3.3\%) & 55.9 & 13.1\\
& 25\%-L & 80.71\% & 75.6 (-2.5\%) & 27.4 & 6.42\\
\hline \hline
\multirow{6}{*}{VWW} & Seed & 85.48\% & 832.4 & 115 & 26.9\\
& 25\%-H & 86.83\% & 206.1 (-0.97\%) & 81.4 & 19.1\\
& 25\%-L & 86.29\% & 208.2 (+0.05\%) & 72.9 & 17.1\\
& 12.5\%-H & 86.30\% & 104.1 (+0.03\%) & 69.5 & 16.3\\
& 12.5\%-L & 81.64\% & 103.8 (-0.2\%) & 34.6 & 8.09\\
& 6.25\%-H & 83.40\% & 50.32 (-3.3\%) & 53.5 & 12.5\\
& 6.25\%-L & 79.36\% & 52.12 (+0.2\%) & 24.5 & 5.73\\
\hline
\end{tabular}
\end{adjustbox}
\end{table}
%
%

Table~\ref{tab:deployment} summarizes the deployment results on the NUCLEO-H743ZI2 for the IC and VWW benchmarks. We do not report KWS results 
%
%
because, as explained above, the structure of the reference CNN makes the trade-offs less interesting.
The table reports two DNNs for each target size (high-OPs ``-H'' and low-OPs ``-L''), corresponding to the two extremes of each Pareto front of Fig.~\ref{fig:pareto_fronts}, neglecting the fully-dominated 50\% front for VWW.
Additionally, for comparison, we also deploy the baseline seed networks.

The Mem. column reports the memory occupation of each model, and the difference in percentage from the imposed constraint. As shown, all networks are within $\pm 3.3\%$ from the target, showing that our constraint formulation produces the expected results.
Further, on the IC task we find solutions with energy consumption spanning from 25.7mJ to 13.4mJ, 20.5mJ to 9.84mJ and 13.1mJ to 6.42mJ respectively for the 75\%, 50\% and 25\% targets.
Noteworthy, the 75\%-H network reduces the energy consumption by 2.2$\times$ with respect to the seed, whit negligible accuracy drop.
Similarly, the deployed solutions for VWW with 25\%, 12.5\% and 6.25\% size consume respectively from 19.1mJ to 17.1mJ, from 16.3mJ to 8.09mJ and from 12.5mJ to 5.73mJ, and the 12.5\%-H CNN reduces the energy consumption of the seed by 1.7$\times$ while improving accuracy of 0.9\%.

\section{Conclusions}
We have proposed a new DNAS formulation that can be used to enhance existing tools allowing them to find DNNs with optimal trade-offs between accuracy and inference complexity, under fixed memory constraints.
With experiments on three different real-world edge-relevant use-cases, we have shown the effectiveness of our method, which is able to reduce the energy consumption by up to 2.2$\times$ with respect to hand-tuned baseline models.

\bibliographystyle{IEEEtran}
\bibliography{bstctl,library}


\end{document}